\documentclass[a4paper,fleqn]{cas-sc}

\usepackage[authoryear]{natbib}

\usepackage{booktabs}
\usepackage{graphicx}
\usepackage{multirow}
\usepackage{xcolor}
\usepackage{amsmath}
\usepackage{ulem}
\usepackage{amsthm}
\usepackage{amssymb}
\usepackage{pifont}
\usepackage[table]{xcolor}
\usepackage{bm}
\usepackage{hyperref}
\usepackage{stfloats}
\usepackage{caption}
\usepackage{varwidth}

\definecolor{darkgreen}{rgb}{0.0, 0.8, 0.0}

\def\tsc#1{\csdef{#1}{\textsc{\lowercase{#1}}\xspace}}
\tsc{WGM}
\tsc{QE}
\begin{document}
\let\WriteBookmarks\relax
\def\floatpagepagefraction{1}
\def\textpagefraction{.001}

% Short title
\shorttitle{Scene-Aware Memory Discrimination}    

% Short author
\shortauthors{Y. Zhong et~al.}  

% Main title of the paper
\title [mode = title]{Scene-Aware Memory Discrimination: Deciding Which Personal Knowledge Stays}  

\author[1]{Yijie Zhong}%[<options>]
% \fnmark[1]
% \ead{dun.haski@gmail.com}

\author[2]{Mengying Guo}

\author[2]{Zewei Wang}

\author[2]{Zhongyang Li}

\author[2]{Dandan Tu}

\author[1]{Haofen Wang}
\cormark[1]
\ead{carter.whfcarter@gmail.com}

\affiliation[1]{
    organization={College of Design and Innovation, Tongji University},
    addressline={281 Fuxin Road},
    city={Shanghai},
    postcode={200092},
    country={China}
}

\affiliation[2]{
    organization={Huawei Technologies Co. Ltd.},
    % addressline={},
    city={Shenzhen},
    postcode={518129},
    country={China}
}

% Footnote text
\fntext[1]{This paper is written during an internship at Huawei 2012 Lab.}
\cortext[cor1]{Corresponding auther}

% For a title note without a number/mark
%\nonumnote{}

% Here goes the abstract
\begin{abstract}
% Here goes the abstract \nocite{*}%% Remove this line from your manuscript.
    Intelligent devices have become deeply integrated into everyday life, generating vast amounts of user interactions that form valuable personal knowledge. Efficient organization of this knowledge in user memory is essential for enabling personalized applications. However, current research on memory writing, management, and reading using large language models (LLMs) faces challenges in filtering irrelevant information and in dealing with rising computational costs. Inspired by the concept of selective attention in the human brain, we introduce a memory discrimination task. To address large-scale interactions and diverse memory standards in this task, we propose a Scene-Aware Memory Discrimination method (SAMD), which comprises two key components: the Gating Unit Module (GUM) and the Cluster Prompting Module (CPM). GUM enhances processing efficiency by filtering out non-memorable interactions and focusing on the salient content most relevant to application demands. CPM establishes adaptive memory standards, guiding LLMs to discern what information should be remembered or discarded. It also analyzes the relationship between user intents and memory contexts to build effective clustering prompts. Comprehensive direct and indirect evaluations demonstrate the effectiveness and generalization of our approach. We independently assess the performance of memory discrimination, showing that SAMD successfully recalls the majority of memorable data and remains robust in dynamic scenarios. Furthermore, when integrated into personalized applications, SAMD significantly enhances both the efficiency and quality of memory construction, leading to better organization of personal knowledge.
\end{abstract}

% Keywords
% Each keyword is seperated by \sep

\maketitle

% Main text
\section{Introduction}

The proliferation of intelligent devices~(\cite{DBLP:journals/informaticaSI/HasanHS24}) has resulted in the continuous collection and storage of vast amounts of personal information generated through daily interactions. Efficient organization and management of these data are essential to support personalized services~(\cite{DBLP:journals/ipm/LiDWLZWL24}). Recent research~(\cite{DBLP:journals/corr/abs-2311-08719,DBLP:journals/tmlr/WangX0MXZFA24,DBLP:conf/acl/LeeHPP023,DBLP:journals/tkde/PibiriT24,DBLP:conf/acl/0001GZW25}) has explored the construction of user memory, which typically involves three key processes: \textbf{memory writing}, to transform collected data into a storable format; \textbf{memory management}, to update and maintain the stored information; and \textbf{memory reading}, to retrieve relevant memories for downstream tasks~(\cite{DBLP:journals/corr/abs-2404-13501}).

However, personal interactions are inherently diverse~(\cite{DBLP:conf/icra/NardelliSR24}), and not all collected data are valuable or necessary to store in user memory. Moreover, these interactions span a wide range of topics, and the relevance of information depends on the specific requirements of applications, which often evolve over time. When memory includes meaningless or irrelevant data, the system may retrieve incorrect historical information or overlook important details during use, potentially leading to erroneous responses. To address this issue, we introduce a \textbf{memory discrimination} task aimed at constructing a more concise and accurate user memory for various applications, such as knowledge graph construction~(\cite{DBLP:journals/ipm/ZhongWGZWW25,DBLP:journals/tkde/ZhangZZZWZ24}), information retrieval~(\cite{DBLP:conf/aaaiss/ToukmajiT24}), and recommendation systems~(\cite{DBLP:journals/ipm/LiDWLZWL24,DBLP:journals/tkde/QuYGZW23}). Inspired by the filter theory of attention(\cite{SelectiveAttention}), which suggests that humans employ an internal mechanism to prioritize relevant information while suppressing distractions, we simulate this cognitive process in our design. Specifically, we conceptualize memory discrimination as a filter during memory construction, formally defined as the process of assessing daily personal interactions according to application requirements to determine whether each piece of data is worth remembering.

Recent research has investigated memory-assisted personalized applications leveraging large language models (LLMs), underscoring their potential in personal memory processing~(\cite{DBLP:journals/corr/abs-2404-13501}). However, incorporating memory discrimination presents two primary challenges: \\
\noindent (1) \textbf{Efficient discrimination at scale}. Real-world interaction data includes a large volume of non-memorable information, such as routine or trivial requests (\textit{e.g.}, operating devices). To construct effective user memory, it is crucial to rapidly and accurately identify valuable interactions. Relying solely on LLMs for this process often results in inefficiency due to their computational overhead. \\
\noindent (2) \textbf{Dynamic definition of memory standards}. LLMs inherently lack an understanding of what information should be remembered. They depend on explicitly defined criteria derived from application needs. These memory standards must remain adaptable to accommodate evolving requirements. Therefore, an effective discrimination method must quickly determine what to remember and what to disregard while precisely communicating these criteria to the model.

Although applications are diverse and complex, making exhaustive enumeration infeasible, the types of personal information they depend on are relatively limited and structurally similar. Prior research~(\cite{DBLP:conf/kes/WestmacottLJ99,DBLP:conf/bioadit/ChoiBLSSY04,DBLP:journals/neuroimage/EtardKBFR19}) indicates that the human brain selectively retains information by focusing on key memory cues. Inspired by Broadbent’s Filter Model~(\cite{SelectiveAttention}), which posits that attention operates by selecting stimuli based on their physical properties, we draw on three fundamental dimensions of personal knowledge (\textbf{personal attributes, relations, and events}) to establish memory scenes that guide the process of memory discrimination.

In this paper, we propose \textbf{SAMD}, a \uline{\textbf{S}}cene-\uline{\textbf{A}}ware \uline{\textbf{M}}emory \uline{\textbf{D}}iscrimination method consisting of a Gating Unit Module (GUM) and Cluster Prompting Module (CPM). The proposed framework identifies memorable interactions and dynamically adapts to the memory requirements of various applications with the support of a frozen LLM. GUM is designed to filter non-memorable data at the early stage of memory processing. It adopts a gating mechanism inspired by selective attention theory~(\cite{AttentionGates}) to help SAMD focus on the most relevant content within each memory scene. A memory scene-based identifier is constructed using salient words that users typically employ in different memory scenes. To enrich the diversity of salient words and reduce the risk of mistakenly filtering meaningful data, we perform multi-view role-playing to simulate diverse user–device interaction contexts. CPM is responsible for defining what should be remembered according to the current application’s requirements and conveying these criteria to the LLM. By integrating user intent signals from sources such as voice assistants and dialogue systems, we analyze large-scale real interaction data to build an intent–scene affinity matrix. Subsequently, we perform intent clustering via matrix decomposition to group similar user intents. For each cluster, memory discrimination rules are constructed based on the associated memory scenes, effectively minimizing redundancy while maintaining complementary coverage across scenes. When an application scenario changes, the identifiers in GUM and discrimination rules in CPM can be easily updated according to the corresponding memory scenes, ensuring fast adaptation to new memory requirements. The main contributions of this paper are summarized as follows:

\begin{itemize}
    \item We investigate the role of memory in personalized applications and introduce a memory discrimination task within the memory construction process. This task aims to identify memorable daily user–device interactions, enabling the construction of high-quality and concise user memory.
    \item We propose SAMD, a Scene-Aware Memory Discrimination method built upon a frozen LLM. SAMD comprises two key modules: GUM, which filters non-memorable data by simulating interactions across memory scenes, and CPM, which defines and communicates memory standards to LLMs by clustering user intents and establishing discrimination rules aligned with application requirements.
    \item We conduct extensive experiments, including both direct and indirect evaluations, to demonstrate the necessity and superiority of SAMD. Results show that SAMD effectively adapts to evolving application demands and remains robust even when user intents are incomplete, incorrect, or unavailable.
\end{itemize}

The remaining sections of this article are organized as follows: Section~\ref{sec:relate} reviews the related work. Section~\ref{sec:problem} formally defines the proposed memory discrimination task, while Section~\ref{sec:method} provides a detailed description of the proposed SAMD method. Comprehensive experimental evaluations are presented in Section~\ref{sec:eval}. In particular, Section~\ref{sec:direct} reports the direct performance of SAMD in the memory discrimination task, and Section~\ref{sec:indirect} presents indirect evaluations of SAMD in the context of memory construction for memory-assisted agents. Finally, Section~\ref{sec:conclusion} concludes the article.

\section{Related Work} 
\label{sec:relate}

This section details the memory construction and personalization mechanisms in LLM-assisted Agents, which are highly relevant to our research. We also discuss how to evaluate the effectiveness of the memory module.

\subsection{Mechanism of Memory Construction}

Attempts have been made to enhance personalized services and applications by personal memory with the help of LLMs~(\cite{DBLP:journals/corr/abs-2404-13501,DBLP:journals/corr/abs-2404-09982,DBLP:journals/corr/abs-2402-11975,DBLP:journals/corr/abs-2406-00057,DBLP:conf/aaai/Wang0S0Y24}). These methods revolve around memory operations like memory writing, memory management, and memory reading. 

% \textbf{Memory Writing.} 
\paragraph{Memory Writing.} TiM~(\cite{DBLP:journals/corr/abs-2311-08719}) extracts the raw information as the relation between two entities and stores them in a structured memory. SCM~(\cite{wang2024enhancinglargelanguagemodel}) retains all historical data and designs a memory controller to decide when and which to activate. MemGPT~(\cite{DBLP:journals/corr/abs-2310-08560}) is entirely self-directed and autonomously updates the memory based on the context. The existing research reflects an important point, which is that the original information is commonly lengthy and noisy. 

\paragraph{Memory Management.} Memorybank~(\cite{DBLP:conf/aaai/ZhongGGYW24}) distills the conversations into a high-level summary and refines the user's knowledge to generate daily insights into personality traits. Voyager~(\cite{DBLP:journals/tmlr/WangX0MXZFA24}) refines the memory based on the feedback of the environment. GITM~(\cite{DBLP:journals/corr/abs-2305-17144}) summarizes the key actions from multiple plans to establish common reference plans for various situations. These methods expend significant resources on removing redundant memory entries or forgetting unimportant memory, whereas a substantial portion of them could have been discarded before being written into memory.

\paragraph{Memory Reading.} This operation extracts related information from memory for reasoning and decision-making. ChatDB~(\cite{DBLP:journals/corr/abs-2306-03901}) generates SQL statements to retrieve historical data from a symbolic memory. MPC~(\cite{DBLP:conf/acl/LeeHPP023}) provides examples for ignoring certain memories when retrieving relevant memories. ExpeL~(\cite{DBLP:conf/aaai/Zhao0XLLH24}) utilizes a vector database to store the memory and obtains the Top-k successful trajectories for the current task. Memory reading and memory writing are collaborative, and the quality of memory greatly influences the methods for memory reading.

\paragraph{} In this paper, we introduce a memory discrimination operation in the memory construction process to determine whether the current context needs to be processed before memory writing based on the application scenario. This mechanism not only helps build high-quality memory but also effectively avoids unnecessary computational and resource costs.

\subsection{Mechanism of Personalization}

Profile-augmented prompting explicitly utilizes summarized user preferences and profiles in natural language to augment LLMs, where prompt engineering acts as a bridge for interaction between users and LLMs~(\cite{DBLP:journals/corr/abs-2502-11528}). Cue-COT~(\cite{DBLP:conf/emnlp/0003WMD0LXW23}) employs chain-of-thought prompting to extract user status like emotion, personality, and psychology. ONCE~(\cite{DBLP:conf/wsdm/LiuCS024}) summarizes topics and user interests based on historical content. These works point out a key technique: Role-Playing in LLMs. It is developed to simulate intricate personas based on various individual profiles and narratives~(\cite{DBLP:journals/corr/abs-2404-18231}). These profiles are constructed using diverse persona data~(\cite{DBLP:journals/nature/ShanahanMR23,DBLP:conf/emnlp/ShaoLDQ23,DBLP:conf/acl/WangPQLZWGGN00024}). The Personas have three categories: individualized persona, character persona, and demographic persona.

\paragraph{Individual Persona} embodies digital clones or personal assistants for individuals~(\cite{DBLP:conf/acl/SalemiMBZ24}) and even mirrors their behaviors~(\cite{DBLP:journals/www/ChenLHWLJPLCWZLC24}). Its applications range from personalized conversation~\cite{DBLP:conf/acl/GaoLZFW23}) and recommendation~(\cite{DBLP:journals/www/ChenLHWLJPLCWZLC24}) to autonomous agents for more complicated task-solving~(\cite{DBLP:journals/corr/abs-2401-05459}). 

\paragraph{Character Persona} is primarily established characters with stories widely recognized by the public. Early research focuses on recognizing characters from a provided text and decodes character traits from their dialogues and actions~(\cite{DBLP:conf/acl/ZhaoZX0Z0024,DBLP:conf/acl/YuLYPZXMZ23}). Recently, researchers have shifted their focus toward applying and prompting LLMs to faithfully reproduce the characters, including their linguistic style~(\cite{DBLP:journals/corr/abs-2308-09597}), knowledge~(\cite{DBLP:conf/acl/WangPQLZWGGN00024}), personality~(\cite{DBLP:conf/acl/WangXHYXGTFL0CL24}), and even decision-making~(\cite{DBLP:journals/corr/abs-2404-12138}).

\paragraph{Demographic Persona} is expected to display unique characteristics of specific groups of people and captures typical traits associated with groups possessing common characteristics~(\cite{DBLP:journals/corr/abs-2306-08158}), such as occupational roles, hobbies or interests, and personality types, \textit{etc.} By assigning specific demographics, LLMs often have better performance in various types of downstream tasks. Demographic personas improve task-solving in both single-agent~(\cite{DBLP:conf/naacl/KongZCLQSZWD24}) and multi-agent~(\cite{DBLP:journals/corr/abs-2311-16542}) systems. Such a persona could also simulate collective social behaviors across various environments~(\cite{DBLP:conf/emnlp/ChawlaWRLG23,DBLP:conf/emnlp/WangCC23,DBLP:journals/corr/abs-2403-13433}).

\paragraph{In this paper,} we establish demographic personas to simulate the interactions of different users with various devices in different scenarios. This ensures that the salient words in GUM are representative and have adequate coverage. Unlike existing works, we do not need to fine-tune the LLMs, nor do we require the LLM to perform role-playing during system usage. SAMD filters out non-memorable data while ensuring the effective recall of memorable data.

\subsection{Evaluation of the Memory}

The evaluation of memory can be broadly categorized into two strategies~(\cite{DBLP:journals/corr/abs-2404-13501}): Indirect Evaluation and Direct Evaluation. Next, we compare their differences in detail and show how we evaluate memory discrimination.

\paragraph{Indirect Evaluation} evaluates the memory module via end-to-end agent tasks. The memory module proves useful if the tasks are effectively accomplished. There are several representative tasks in the recent works. Engaging in human-AI conversations is a key application of agents, where memory significantly enhances user experience. MemoChat~(\cite{DBLP:journals/corr/abs-2308-08239}) evaluates the agents on interactive dialogues using GPT-4 to score the responses. Research utilizes the CSIM~(\cite{DBLP:journals/corr/abs-2310-08560}) score and SCE-p~(\cite{DBLP:conf/acl/LeeHPP023}) score to evaluate the memory effect on increasing engagement of users. Question answering can evaluate the memorized information. ReAct~(\cite{DBLP:conf/iclr/YaoZYDSN023}) evaluates the memory from inside-trial information and external information from Wikipedia. Retroformer~(\cite{DBLP:conf/iclr/YaoHNLFXNC0AXMW24}) and Reflexion~(\cite{DBLP:conf/nips/ShinnCGNY23}) further include the cross-trial information. The success rate refers to the proportion of tasks that agents can successfully solve. Expel~(\cite{DBLP:conf/aaai/Zhao0XLLH24}) accesses how many special tasks can be correctly completed through memory in AlfWorld, while GITM~(\cite{DBLP:journals/corr/abs-2305-17144}) evaluates this in Minecraft. In addition, some methods~(\cite{DBLP:journals/tmlr/WangX0MXZFA24}) use the exploration degree in exploratory games to reflect the extent that agents can explore the environment.

\paragraph{Direct Evaluation} independently measures the capability of the memory module. Previous studies apply both subjective and objective metrics. Memorybank~(\cite{DBLP:conf/aaai/ZhongGGYW24}) and TiM~(\cite{DBLP:journals/corr/abs-2311-08719}) recruit human evaluators to check if the memory contains reasonable answers for the current question and if the recalled memory is suitable for the current context. Other works~(\cite{wang2024enhancinglargelanguagemodel,DBLP:conf/acl/LeeHPP023}) focus on the relation and contradiction between the current query and historical memory. Numeric metrics are also used to evaluate the effectiveness and efficiency of the memory module. The correctness is calculated to evaluate whether the recalled memory~(\cite{DBLP:journals/corr/abs-2306-03901}) and the responses~(\cite{DBLP:journals/corr/abs-2310-08560}) from the agents match the correct answers. A reference accuracy is used to assess the retrieval process of the memory by calculating the F1-score~(\cite{DBLP:conf/aaai/ZhongGGYW24,DBLP:journals/corr/abs-2308-08239}). The time and hardware cost~(\cite{DBLP:conf/nips/Tack0MSTS24}) is widely used to assess the efficiency of memory operations.

\paragraph{In this paper,} to comprehensively validate the necessity of memory discrimination and the effectiveness of SAMD, we not only calculated indirect metrics, such as question answering and historical memory retrieval accuracy, but also analyzed direct metrics like the accuracy and recall of memorable and non-memorable data. We also calculate the computational cost to reflect the efficiency of our method.

\begin{figure}[pos=ht]
    \centering
    \includegraphics[width=\linewidth]{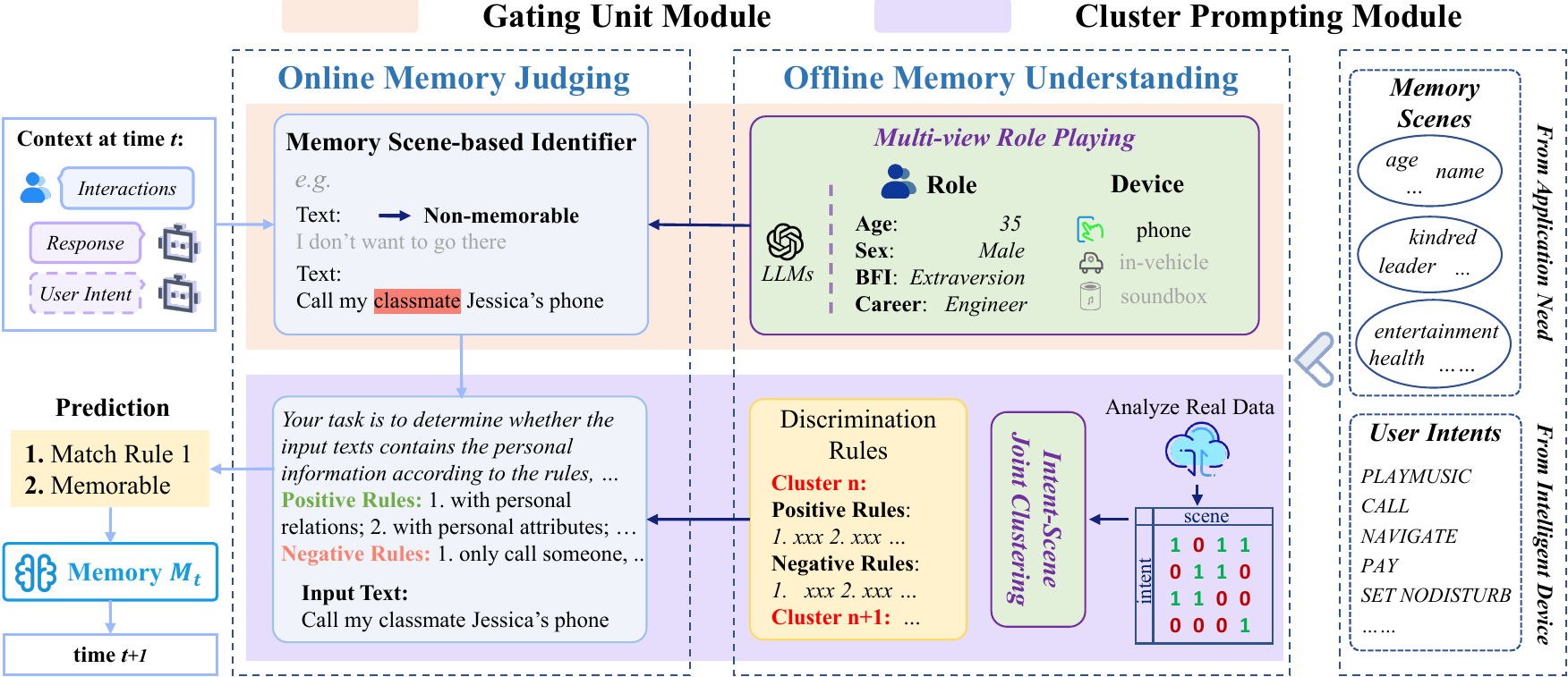}
    \caption{The overview of the proposed memory discrimination method via the gating unit module and cluster prompting modules. It has two phases: offline memory understanding and online memory judging. In the memory understanding phase, a memory scene-based identifier and cluster prompts with discrimination rules are created using all memory scenes and user intents. During the memory judging phase, each sentence in the context at time $t$ is evaluated for whether to remember, with memorable sentences being stored in memory $M_t$ for use in time $t+1$.}
    \label{fig:pipeline}
\end{figure}

\section{Problem Definition}
\label{sec:problem}

Constructing user memory is crucial for memory-assisted LLM-based personalized applications. The memory discrimination problem is defined as a binary classification task to build more efficient and accurate user memory.
This could be described as:

\noindent \textbf{Memory Discrimination ($\textbf{D}$)}: Judges whether the sentences in context should be recorded in memory based on the demand of personal information.

Consider that at time $t$, the context $C_t$ consists of user interactions $q_t$, response $o_t$, and user intent $I=\{I_i\}^n_{i=1}$ of a dialogue system, question answering system, or voice assistant.
The memory discrimination method $\mathcal{F}$ determines whether each sentence $l_i$ in $C_t$ should be recorded (\textit{memorable}) as memory $M_t$ or discarded (\textit{non-memorable}), with the help of additional information in $C_t$, such as user intents. This can be formalized as follows: 
\begin{equation}
    \mathcal{F}_\phi(l_i, I) \rightarrow \{0, 1\},
\end{equation}
where $I$ is optional, and $\phi$ denotes the memory criterion formed by the application's needs for personal information. These needs can be described as a series of memory scenes $S=\{S_i\}^m_{i=1}$ linked to personal information, which includes elements related to personal relations, attributes, and events.

Based on the results of $\mathcal{F}$, we recompile all the memorable data with predicted results of 1 to obtain $C^{'}_t$ and store it in memory. Therefore, $\mathcal{D}$ can be implemented using function $\mathcal{F}$.

Previous research~(\cite{DBLP:journals/corr/abs-2404-13501}) represents memory-assisted LLM-based personalized applications as:
\begin{equation}
    o_{t+1} = \text{LLM}\{ \mathbf{R}( \underbrace{\mathbf{P}(M_{t-1}, \mathbf{W}(C_t))}_{M_t}, q_{t+1}) \}.
\end{equation}
Existing work focuses on the following three core operations related to memory:

\begin{itemize}
    \item \textbf{Memory Writing ($\mathbf{W}$)}: Projects the raw context $C_t$ into the actually stored memory contents. 
    \item \textbf{Memory Management ($\mathbf{P}$)}: Processes the stored memory to make it more effective through summarization, integration, or forgetting mechanisms.
    \item \textbf{Memory Reading ($\textbf{R}$)}: Obtains relevant information from memory $M_t$ to support the next user interactions $q_{t+1}$ and produce the response $o_{t+1}$.
\end{itemize}

However, due to the vast number of user interactions and the abundance of non-memorable context, processing all raw data indiscriminately can lead to significant unnecessary time and computational costs. Moreover, noise data in memory can lead to erroneous retrieval results and incorrect responses. Therefore, we propose adding a memory discrimination operation, which transforms the construction of memory $M_t$ into:
\begin{equation}
    M_t = \mathbf{P}(M_{t-1}, \mathbf{W}(\underline{\mathbf{D}(C_t)})).
\end{equation}

\section{Proposed Method}
\label{sec:method}

\subsection{Overview}

\autoref{fig:pipeline} illustrates the proposed SAMD, comprising the gating unit module (GUM) and cluster prompting module (CPM). 
SAMD distinguishes and retains memories through offline memory understanding and online memory judging. 
Offline, we convey to SAMD what to remember and clarify the memory scenes based on application needs.
GUM constructs a memory scene-based identifier through multi-view role-playing, while CPM creates cluster prompts by examining the relationship between intents and scenes and a joint clustering.
Online, GUM verifies incoming user interactions using the identifier.
A failed verification indicates the data is non-memorable. Otherwise, CPM employs LLMs and cluster prompts to ascertain if this data needs to be remembered.

\begin{figure}[pos=t]
    \centering
    \includegraphics[width=\linewidth]{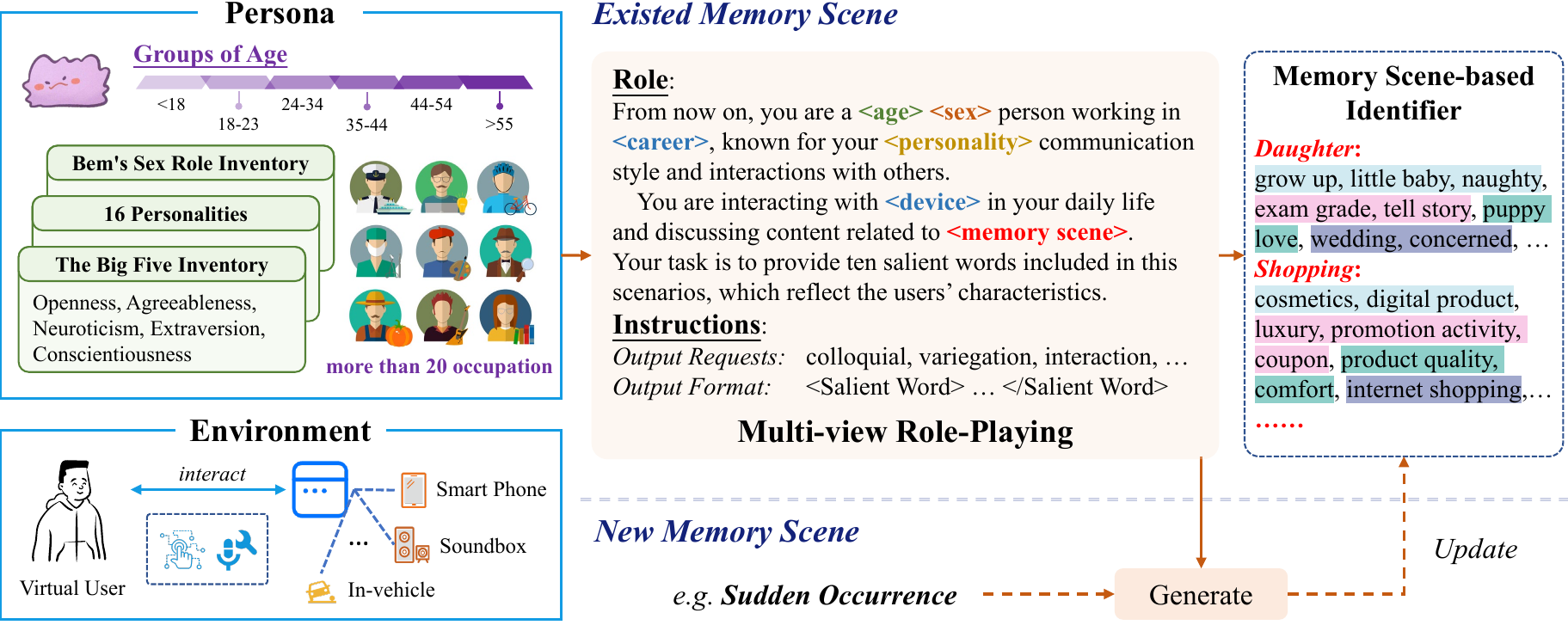}
    \caption{Instructions for constructing the memory scene-based identifier in GUM, enabling quick updates for new memory scene demands. Different individuals focus on various aspects of the same memory scene.}
    \label{fig:gate}
\end{figure}

\subsection{Gating Unit Module}

Users interact with numerous intelligent devices daily, yet only a fraction of these interactions are memorable. To avoid inefficiently processing all data with LLMs, we introduce a gating mechanism inspired by human selective attention when facing abundant information from the environment~(\cite{AttentionGates}).
Specifically, we propose the gating unit module to filter out non-memorable data and avoid unnecessary computational costs on these data while maintaining the recall of the memorable data. The structure of GUM is detailed in \autoref{fig:gate}.

Relying on sentence-level embedding similarity comparisons between current data and memory scenes is computationally expensive and inefficient. Moreover, it may retain non-memorable data due to ambiguous semantic matching. Therefore, we propose a word-level gating approach.
Following Broadbent’s Filter Model~(\cite{SelectiveAttention}), which suggests selecting which stimuli to process based on their physical properties, we construct a memory scene-based identifier to prioritize relevant data while filtering out the rest.

\paragraph{Memory Scene-based Identifier.} We propose to use salient words that frequently occur in different memory scenes for efficient filtering. 
To ensure a high recall of memorable data, we need to capture fully the focal points and expression styles of different users.
We utilize LLMs to simulate user-device interactions in memory scenes $S$ through multi-view role-playing. 
We define users $\mathcal{R}$ with diverse attributes: \textit{age, sex, career, personality}. The salient words list $W_i$ created by a specific user in memory scene $S_i$ can be formally defined as:
\begin{equation}
    W_i = \bigcup \mathcal{L}_\phi(S_i|\mathcal{R}),
\end{equation}
where $\mathcal{L}_\phi$ denotes the employed LLM. The memory scene-based identifier can be obtained by merging all the sets ($K_i$). In the memory judging phase, salient words in the identifier are matched with the input text, and upon a successful match, the text is forwarded to CPM.

\paragraph{Handle New Memory Scene.} In practical use, new memory scene $S_{m+1}$ may arise, as shown by `\textit{sudden occurrence}' in Figure~\ref{fig:gate}. GUM provides excellent scalability via multi-view role-playing, enabling quick acquisition of the corresponding salient words $W_{m+1}$ and updating the identifier. 

\begin{figure}[pos=t]
    \centering
    \includegraphics[width=\linewidth]{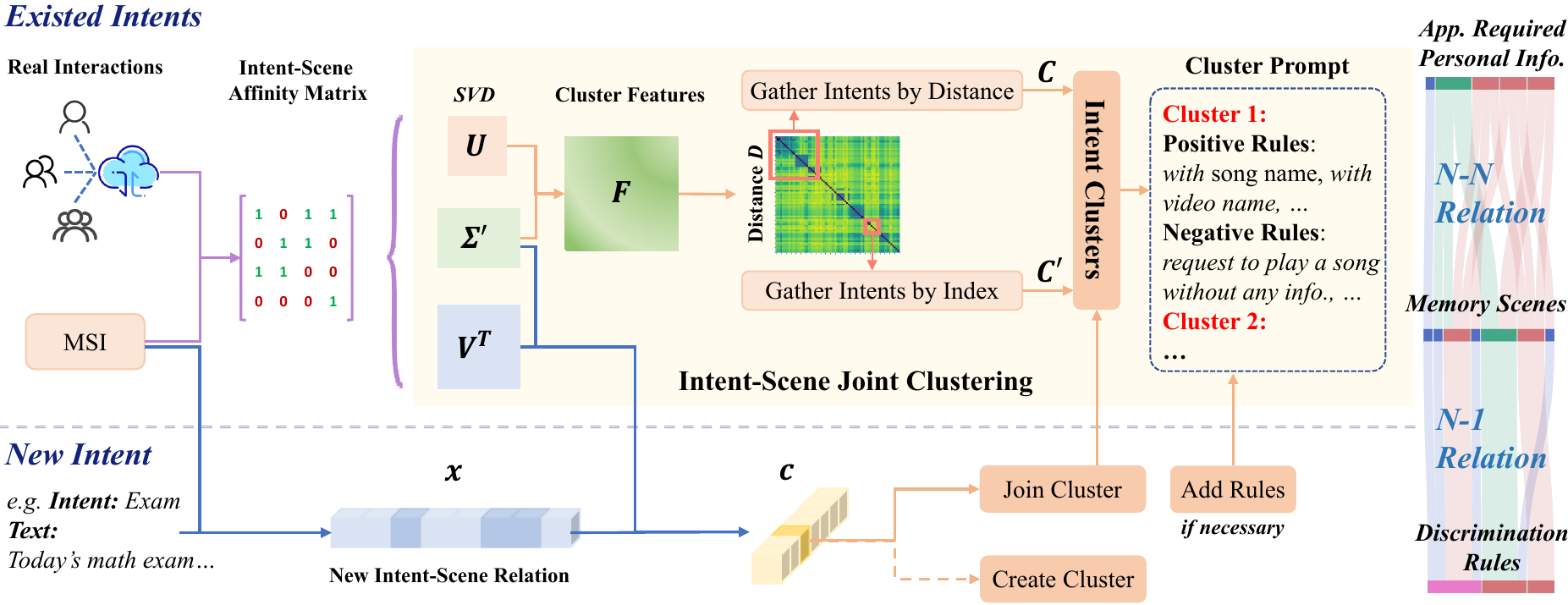}
    \caption{The left side illustrates the proposed Cluster Prompting Module. MSI represents the memory scene-based identifier in GUM. Intents are grouped into different clusters via matrix decomposition. The cluster prompts consist of several positive and negative rules. The rightmost side demonstrates how to design discrimination rules using memory scenes as a bridge to represent the personal information required by real-world applications.}
    \label{fig:cluster}
\end{figure}

\subsection{Cluster Prompting Module}

Instructing LLMs on what to remember and what not to remember assists them in making accurate memory discrimination. Summarizing an application's memory requirements using only a few general descriptions can lead to cognitive biases in LLMs. On the other hand, providing detailed descriptions for each memory scene may involve excessive manual effort, increase computational costs, and reduce scalability.
Thus, we analyze the relationship between intents and memory scenes on 80 million real data, creating an intent-scene affinity matrix $E\in\{0,1\}^{n\times m}$. Analysis reveals that some intents share similar memory scenes. As shown in \autoref{fig:cluster}, we propose to cluster the intents and build discrimination rules for each cluster based on its associated memory scenes.

\paragraph{Intent-Scene Joint Clustering.} Considering intents are categorized into $K$ classes, $\{A_k\}^K_{k=1}$, and memory scenes into $Q$ classes, $\{B_q\}^Q_{q=1}$. The relationship between intent $I_j$ and memory scene $S_i$ can be expressed as:
\begin{equation}
    P(S_i|I_j) = \sum\nolimits_{k,q}P(S_i|B_q)P(B_q|A_k)P(A_k|I_j).
\end{equation}
Thus, we have $P(S|I) = P(A|I)P(B|A)P(S|B)$. The intent clustering can be obtained through $P(A|I)$ if we consider $E\sim P(S|I)$. 
Researches show that matrix factorization, like singular value decomposition (SVD)~(\cite{szalontai2020svd}), can efficiently generate effective representations~(\cite{DBLP:conf/wsdm/QiuDMLWT18,DBLP:conf/www/QiuDMLWWT19}) or clustering insights~(\cite{DBLP:conf/aaai/GaoXWXZ20}). As it can yield reasonable cluster centroids and cluster membership indicators~(\cite{DBLP:conf/iccv/ZassS05,DBLP:conf/sdm/DingH05}), we apply SVD to generate approximate intent clustering. We get $E=U\Sigma V^T$, and select a subset of singular values $\Sigma'$, 
ensuring the ratio $\frac{|E'_{i,j}>\sigma|}{|E_{i,j}=1|}$ exceeds 0.9 after reconstruction, with $\sigma$ as a threshold. 
Highly similar intents are grouped based on their distances $D=F\cdot F^T$ from each other. Thus we obtain clusters $C=[C_1, C_2, \cdots]$ based on the criteria: for every intent $I_u,I_v\in C_i$, it need to be satisfied
\begin{equation}
    d(I_u, I_v) = D_{u, v} \leq \min(\text{median}(D_u), \epsilon),
\end{equation}
where $d(\cdot,\cdot)$ means the distance between intents, $\text{median}(\cdot)$ denotes the median of the list, and $\epsilon$ signifies a minimal distance. We only keep clusters with more than two intents.
For the remaining intents, we extract clustering information from $F$ and select $\text{Index}_i=\operatorname{argmax}_j F_{i,j}$. We group intents with the same $\text{Index}$ into the same cluster to obtain $C'$:
\begin{equation}
    C' = [(I_u,I_v,\cdots|\text{Index}_u=\text{Index}_v),\cdots].
\end{equation}
Finally, we merge $C$ and $C'$ as the intent clustering, followed by manually verifying and merging any two clusters that cover the same memory scenes while reassessing the independent intent.

Each cluster prompt includes discrimination rules based on related memory scenes: positive rules indicate what to remember and negative rules for not to remember. A text is non-memorable if it fails to meet any positive rules or meets negative ones. LLMs follow a two-step process. They first think and provide the evidence based on these rules, then present the discrimination result.

\paragraph{Handle New Intent.} When the system supports a new intent, CPM can integrate it with limited cost. We obtain the relative vector $x\in\{0,1\}^{1\times m}$ using the memory scene-based identifier with the intent and its corresponding interactions. The cluster feature for the new intent can be calculated as $c=\sqrt{\Sigma'}\cdot V^T\cdot x^T$. We identify the cluster by finding the maximum value in $\mathtt{softmax}(c)$. After verification, we decide whether to add it to the existing cluster or create a new one, updating the relevant discrimination rules as necessary.

\paragraph{Handle Incorrect or Unavailable Intent.} SAMD does not need the device to identify user intents, thanks to this module's error-tolerance mechanism. Cluster prompting ensures that similar intents with consistent memory scenes share discrimination rules, reducing the impact of intent deviations. When user intents are unavailable, we apply the memory scene-based identifier to match input text and utilize the intent-scene affinity matrix to obtain candidate intents.

\section{Evaluation}
\label{sec:eval}

In this section, to comprehensively evaluate the superiority of our method and the value of memory discrimination in the memory construction process, we conduct direct evaluations of the memory discrimination module and indirect evaluations of the LLM-assisted agents integrated with this module. 

\subsection{Direct Evaluation of Memory Discrimination}
\label{sec:direct}

For direct evaluation, we focus on the performance of SAMD to determine whether user interactions are memorable.

\subsubsection{Experimental Settings}

\paragraph{Dataset.}
We incorporate a broader range of data from personal daily interactions, including both voice and text-based interactions with intelligent assistants on devices such as cars, smart speakers, and mobile phones. In total, we first collect approximately 80 million anonymous interaction records from real users (with informed consent), encompassing over 2,600 distinct intents recognized by the devices. From these, we select 97 high-value intents based on their relevance to personal memory construction. After data cleaning and filtering, we construct multiple datasets for direct evaluation of the proposed method. To comprehensively evaluate memory discrimination under different conditions, we construct two real-world datasets, one synthetic dataset, and two datasets targeting long-tail distributions. Detailed dataset statistics are provided in \autoref{tab:datasets}.

\begin{table}[pos=b]
\caption{Statistics of the datasets under different settings to directly evaluate memory discrimination. `Num.' means the number.}
\centering
\resizebox{5in}{!}{
\begin{tabular}{@{}l|c|c|ccc@{}}
\toprule
\multirow{2}{*}{\textbf{Dataset}} & \multirow{2}{*}{\textbf{Num. of Memory Scenes}} & \multirow{2}{*}{\textbf{User Intents}} & \multicolumn{3}{c}{\textbf{Num. of Data}} \\
 &  &  & \textbf{Total} & \textbf{Memorable} & \textbf{Non-memorable} \\ \midrule
\textbf{BID-20K} & 203 & \checkmark & 19,416 & 10,567 (54.4\%) & 8,849 (45.6\%) \\
\textbf{IID-10K} & 203 & \checkmark & 9,700 & 1,008 (10.4\%) & 8,692 (89.6\%) \\
\textbf{GID-1K} & 24 & \ding{55} & 1,000 & 500 (50\%) & 500 (50\%) \\ \midrule 
\textbf{LID} & 10 & \checkmark & 500 & 275 (57.0\%) & 215 (43.0\%) \\ 
\textbf{MLID} & 213 (203+10) & \checkmark & 2500 & 1330 (53.2\%) & 1170 (46.8\%) \\ 
\bottomrule
\end{tabular}}
\label{tab:datasets}
\end{table}

\begin{itemize}
    \item \textbf{Balanced interaction dataset (BID-20K)}: This dataset includes more than 100 samples per intent, ensuring that the number of memorable and non-memorable samples is approximately balanced for fair evaluation.
    \item \textbf{Imbalanced interaction dataset (IID-10K)}: Each intent contains 100 samples sampled according to the real-world distribution of memorable and non-memorable data, simulating realistic usage scenarios.
    \item \textbf{Generated interaction dataset (GID-1K)}: This synthetic dataset simulates dynamic scenarios where intents and memory scenes evolve. We first create 10 virtual users and generate personal details for their family members, classmates, friends, and acquaintances (100 individuals in total). An event list with 500 events is then constructed based on these entities and their associations across different memory scenes. Finally, we generate interaction data according to these events to model diverse and evolving user behaviors.
    \item \textbf{Long-tail interaction dataset (LID)}: To evaluate performance under long-tail distribution conditions in reality, we identify memory scenes whose interaction frequency is lower than 1\% of the average across all existing scenes. This dataset contains 10 rare memory scenes with 500 samples. Representative long-tail scenarios include niche domains (\textit{e.g.}, ingredients, collections, pets, niche topics) and cross-domain compositions (\textit{e.g.}, work and special diet, travel and customs).
    \item \textbf{Merged long-tail interaction dataset (MLID)}: This dataset merges 203 regular memory scenes with the 10 identified long-tail memory scenes, forming a total of 213 scenes with 2,500 samples. It is used to assess model robustness and adaptability when rare and frequent interactions coexist.
\end{itemize}

Data is first annotated by three annotators, treating all memory scenes as the application demands. To avoid erroneous annotations caused by inconsistent labeling, we adopt a widely recognized annotation alignment mechanism. When annotations are consistent, it is determined whether a data point is memorable. When disagreements arise among annotators, another annotator is introduced to re-annotate the data. The team then discusses to reach a consensus on the final annotation. If no agreement can be achieved, the data will be discarded.

\paragraph{Metrics.} For memory discrimination, following the previous research~(\cite{DBLP:journals/corr/abs-2404-13501,wang2024enhancinglargelanguagemodel}), we apply widely used metrics: Accuracy (Acc) and weighted-F1 score (\bm{$wF_1$})~(\cite{DBLP:conf/aaai/WangWPY17}). We also report the Recall of memorable (\bm{$R_1$}) and non-memorable  (\bm{$R_0$}) data to determine if LLMs understand what to remember and what not to remember. For intent clustering, following the previous works~(\cite{DBLP:journals/corr/abs-2305-07280,DBLP:journals/tkde/FangLLGJZ23}), we use the Rand Index (RI)\footnote{https://en.wikipedia.org/wiki/Rand\_index}, which measures the similarity between two clusterings. The value of RI ranges from 0 to 1, with larger values indicating better performance.

\paragraph{Compared Methods.} For memory discrimination, we select different types of approaches in the memory-assisted agent that focus on memory writing function, which is directly linked to this task: SCM~(\cite{wang2024enhancinglargelanguagemodel}), Memochat~(\cite{DBLP:journals/corr/abs-2308-08239}), and TiM~(\cite{DBLP:journals/corr/abs-2311-08719}). For intent clustering, we compare our method with the approach of directly clustering embeddings. The embeddings are extracted from BERT~\cite{DBLP:conf/naacl/DevlinCLT19} and GPT-2~(\cite{radford2019language}). Then, we utilize several density-based clustering algrithms~(\cite{DBLP:journals/corr/abs-2306-09256,DBLP:journals/tkde/LiZZC24}), which can adaptively form multiple clusters, including MeanShift, Bisecting K-Means, and Spectral\footnote{https://scikit-learn.org/stable/modules/clustering.html}.

\paragraph{LLMs.} We employ different series of LLMs with varying scales: MiniCPM-2B~(\cite{DBLP:journals/corr/abs-2404-06395}), Qwen2~(\cite{qwen2}), LLAMA3~(\cite{DBLP:journals/corr/abs-2302-13971}), and GLM series~(\cite{DBLP:journals/corr/abs-2406-12793}). This helps verify if SAMD adapts to different LLMs and performs well on LLMs with few parameters.

\paragraph{Implementation Details.} The LLMs used in our experiments are all deployed offline to protect data privacy. Based on 203 memory scenes, the memory scene-based identifier contains 31,831 salient words. The cluster prompting module consists of 16 clusters with 91 discrimination rules.

\begin{figure}[pos=t]
    \centering
    \includegraphics[width=\linewidth]{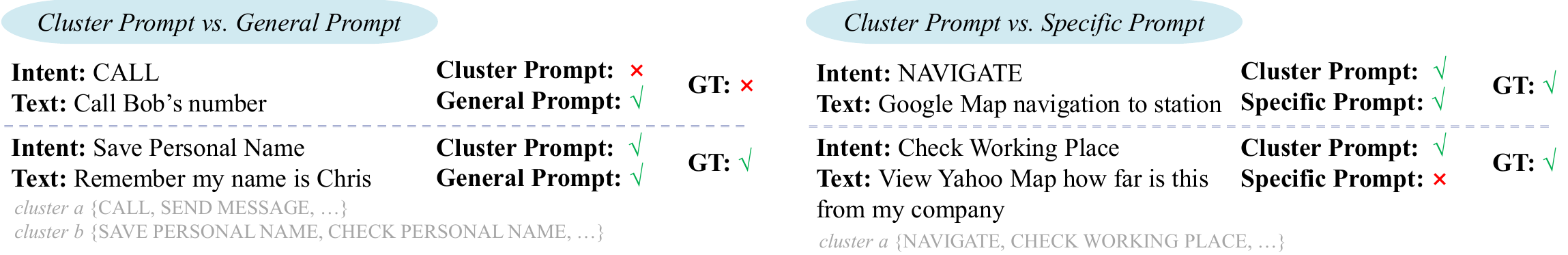}
    \caption{Comparative examples of various prompts.}
    \label{fig:case}
\end{figure}

\subsubsection{Main Results}

We evaluate the superiority, generalization, robustness, and efficiency of our method, respectively.

\paragraph{Superiority of proposed SAMD.} As shown in \autoref{tab:total}, the proposed SAMD (with intent) is at least 3$\times$ faster while providing over 20\% greater accuracy in memory discrimination. It aids LLMs in understanding what to remember, retaining more than 90\% of the memorable data, and improving overall performance by over 20\%.

\noindent
\begin{minipage}[pos=t]{\textwidth}
    \begin{minipage}[pos=t]{0.55\textwidth}
        \captionsetup{type=table,width=\textwidth}
        \caption{The quantitative results comparing existing methods under different settings. $\mathtt{Qwen2}$ and $\mathtt{LLAMA3}$ represents Qwen2-7B and LLAMA3-8B respectively. `\textit{w/o} GUM' indicates that GUM is removed in the experiments.}
        \resizebox{\textwidth}{!}{
            \begin{tabular}{@{}cccccccccc@{}}
\toprule
\multicolumn{1}{c|}{\multirow{2}{*}{\textbf{Method}}} & \multicolumn{3}{c|}{\textbf{BID-20K}}                          & \multicolumn{3}{c|}{\textbf{IID-10K}}                          & \multicolumn{3}{c}{\textbf{GID-1K}}       \\
\multicolumn{1}{c|}{}                                 & \textbf{Acc} & \bm{$R_1$} & \multicolumn{1}{c|}{\textbf{Time}} & \textbf{Acc} & \bm{$R_1$} & \multicolumn{1}{c|}{\textbf{Time}} & \textbf{Acc} & \bm{$R_1$} & \textbf{Time} \\ \midrule
\multicolumn{1}{c|}{\textbf{SCM}}                     & 54.4         & -          & \multicolumn{1}{c|}{-}             & 10.4         & -          & \multicolumn{1}{c|}{-}             & 50.0         & -          & -             \\
\multicolumn{1}{c|}{\textbf{Memochat}}                & 65.6         & 68.4       & \multicolumn{1}{c|}{1.39}          & 16.3         & 73.1       & \multicolumn{1}{c|}{1.21}          & 52.9         & 76.4       & 1.72          \\
\multicolumn{1}{c|}{\textbf{TiM}}                     & 48.6         & 50.2       & \multicolumn{1}{c|}{1.26}          & 67.2         & 50.9       & \multicolumn{1}{c|}{0.95}          & 63.7         & 68.0       & 1.60          \\ \midrule
\multicolumn{10}{c}{\textit{with Intent}}                                                                                                                                                                                           \\ \midrule
\multicolumn{1}{l|}{\textbf{Ours$_\mathtt{Qwen2}$}}   & 85.4         & 89.5       & \multicolumn{1}{c|}{0.45}          & 82.5         & 93.6       & \multicolumn{1}{c|}{0.18}          & -            & -          & -             \\
\multicolumn{1}{l|}{\textit{w/o} GUM}                            & 80.9         & 90.9       & \multicolumn{1}{c|}{0.55}          & 77.6         & 96.2       & \multicolumn{1}{c|}{0.53}          & -            & -          & -             \\
\multicolumn{1}{l|}{\textbf{Ours$_\mathtt{LLAMA3}$}}  & 85.1         & 92.1       & \multicolumn{1}{c|}{0.43}          & 83.0         & 95.7       & \multicolumn{1}{c|}{0.18}          & -            & -          & -             \\
\multicolumn{1}{l|}{\textit{w/o} GUM}                            & 80.5         & 92.3       & \multicolumn{1}{c|}{0.54}          & 81.1         & 97.2       & \multicolumn{1}{c|}{0.53}          & -            & -          & -             \\ \midrule
\multicolumn{10}{c}{\textit{without Intent}}                                                                                                                                                                                        \\ \midrule
\multicolumn{1}{l|}{\textbf{Ours$_\mathtt{Qwen2}$}}   & 80.4         & 99.6       & \multicolumn{1}{c|}{0.98}          & 80.0         & 96.8       & \multicolumn{1}{c|}{0.20}          & 76.4         & 97.9       & 0.94          \\
\multicolumn{1}{l|}{\textbf{Ours$_\mathtt{LLAMA3}$}}  & 86.0         & 97.2       & \multicolumn{1}{c|}{0.96}          & 84.1         & 92.5       & \multicolumn{1}{c|}{0.28}          & 80.2         & 95.6       & 0.89          \\ \bottomrule
\end{tabular}
        }
        \label{tab:total}
    \end{minipage}\hfill
    \begin{varwidth}[pos=t]{0.4\textwidth}
        \vspace{0.2cm}
        \includegraphics[width=\textwidth]{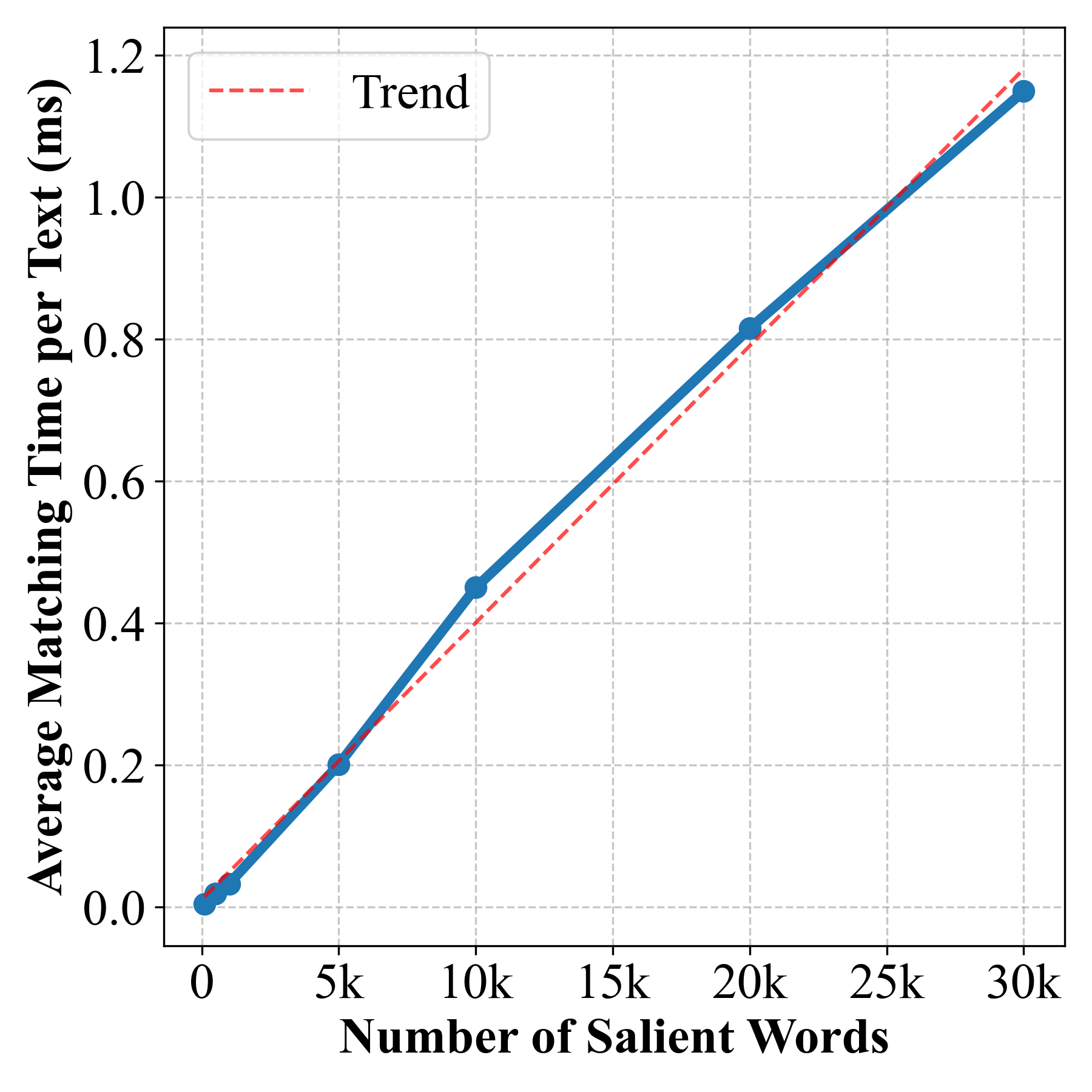}
        \captionsetup{type=figure}
        \vspace{-0.3cm}
        \caption{Efficiency of Salient Words Matching.}
        \label{fig:matching}
    \end{varwidth}
\end{minipage}

\paragraph{Generalization of proposed SAMD.} Results across different datasets in \autoref{tab:total} demonstrate that SAMD performs well across various conditions, including dynamically changing memory scenes and scenarios without explicit user intent. When user intent is absent, SAMD still achieves high discrimination accuracy with only a minor increase in processing time, maintaining a clear speed advantage over existing methods.

From the perspective of diverse memory scenes, examples shown in \autoref{fig:case} illustrate SAMD’s ability to flexibly adjust to diverse contexts. When comparing cluster prompts with general prompts, it is evident that the memory standard for identical content (e.g., a person’s name) may vary between texts. Discrimination rules in general prompts often struggle to accommodate such variability. Similarly, when compared with specific prompts, the latter frequently overlook certain intents due to the broad range of memory scenes and heterogeneous user expressions. In contrast, cluster prompt groups semantically related intents, enabling the associated discrimination rules to cooperate, thereby minimizing the possibility of missing memorable texts and enhancing adaptability across different memory scenes.

From the perspective of long-tail memory scenes, SAMD also exhibits strong generalization. On the LID, achieves an accuracy of 87\% and a weighted-F1 score of 86\%, demonstrating that salient words generated through multi-view role-playing effectively capture the distinct linguistic characteristics of rare or infrequent memory scenes. Similarly, the discriminative rules within CPM support efficient adaptation to low-frequency interaction contexts. On the MLID, SAMD attains an accuracy of 85\% and a weighted-F1 score of 85\%, confirming that the cluster prompts successfully distinguish regular scenes from long-tail ones, reducing the risk of misidentification for infrequent data.

\paragraph{Robustness of proposed SAMD.} We validate through experiments involving intent recognition uncertainty and analyses of potential LLM hallucination effects.

From the perspective of intent uncertainty, we simulate scenarios in which user intents are incomplete or noisy rather than explicitly incorrect. In this experiment, the accurate intent labels in BID are deliberately removed, and SAMD is required to infer candidate intents using its internal reasoning mechanism, which integrates the memory scene-based identifier with the intent–scene affinity matrix. The results show that the ground-truth intents appear among the generated candidate intents in 70.26\% of all cases, and that the ground-truth and inferred intents fall within the same cluster in 89.79\% of the instances. This confirms that CPM provides strong robustness to SAMD: even when the inferred intents contain noise, their cluster proximity allows accurate memory discrimination by leveraging semantically related memory scenes to approximate users’ original intents.

From the perspective of potential hallucination effects, we systematically assess whether the frozen LLM exhibits hallucinations in our task. Specifically, the LLM is instructed to produce explicit reasons for each prediction, detailing the rules applied (positive or negative) and the supporting textual evidence. We manually examine 3,000 erroneous predictions and categorize hallucinations into two types: (1) \textbf{Fabricated Content}, where key entities or attributes cited in the explanation are absent from the source interaction text; (2) \textbf{Rule Inconsistency}, where the reason contradicts the final predicted label (\textit{e.g.}, a positive rule cited for a negative result). we identify 135 hallucinated cases, corresponding to a hallucination rate below 5\%. Among these, 26 cases (less than 1\% of all errors) involve fabricated content, while 109 cases correspond to rule inconsistencies. These results indicate that LLM hallucinations primarily manifest as mild reasoning inconsistencies rather than invented content, and can be further mitigated by more explicit prompting or structured instructions. Overall, the low hallucination rate demonstrates that LLM reasoning errors have a negligible impact on SAMD’s memory discrimination reliability.

We further evaluate SAMD under adversarial annotation errors and noisy rule interference. To assess robustness against adversarial label perturbations, we simulate label-flipping scenarios in which user interactions related to `preferred music genre' and `frequently used music player' are originally labeled as memorable, and the labels of one category (\textit{e.g.}, data related to `music players') are flipped from memorable to non-memorable. To accommodate this adversarial setting, the corresponding discrimination rules are simultaneously adjusted by transferring a subset of positive rules into negative rules. Results show that, within the affected intents, SAMD still achieves an accuracy of 86.7\%, demonstrating strong resilience to adversarial label noise.
We then evaluate robustness to irrelevant rule noise by generating and injecting 10 discrimination rules unrelated to existing intents, including 5 positive and 5 negative rules. These noise rules are randomly inserted into existing intents, and experiments are conducted on the GID dataset. The results indicate that $R_1$ of SAMD remains largely unaffected, reaching 96.8\%, which confirms its robustness against rule-level noise. Furthermore, when requiring the LLMs to explicitly report the rules they relied upon during judgment, we observe that fewer than 1\% of the outputs reference the injected noise rules. This observation further verifies the high robustness of the proposed cluster prompt.

\paragraph{Efficiency of proposed SAMD.} First, we verify that the GUM module maintains high matching efficiency even as the number of salient words increases substantially through both empirical analysis and theoretical estimation. Experiments are conducted using 20,000 samples on a 13th Gen Intel(R) Core(TM) i7-13700K CPU. The results indicate an average throughput of approximately 25 million salient-word matches per second. In practical conditions where user intent information is available, the average matching time per text is only 0.08ms, which is computationally negligible. As shown in \autoref{fig:matching}, even when matching against all 31,831 salient words, the average processing time per text remains around 1ms. Extrapolation suggests that if both the number of memory scenes and salient words are to expand by three orders of magnitude, the average per-text matching time would still remain below one second, demonstrating the strong scalability of the SAMD design. It is also worth noting that a larger set of salient words is not inherently advantageous. Salient words should be representative and discriminative rather than merely abundant. Excessive expansion introduces redundant or weakly informative terms, offering diminishing returns. These findings confirm that GUM effectively handles salient-word list growth while preserving practical efficiency.

Second, detailed measurements are conducted across all stages of SAMD’s offline memory understanding and updating phases. Constructing the memory scene-based identifier for 203 memory scenes with 31,831 salient words requires 22 minutes, consuming 2.5M input tokens, 0.6M output tokens, and incurring a total cost of only \$0.49. The intent clustering step executes extremely rapidly, completing within 0.04 seconds. Incremental updates are also efficient: adding a new memory scene takes approximately 6 seconds and requires 16K total tokens, while adding a new intent, using 1,000 corresponding interactions as examples, requires only 1.2 seconds. These results demonstrate that the offline phase introduces minimal computational overhead and that incremental updates for new memory scenes or intents can be performed within seconds. Overall, SAMD exhibits excellent computational efficiency, scalable processing capability, and low maintenance cost, making it suitable for deployment in large-scale, continuously evolving user–device interaction environments.

\begin{table}[t]
\caption{Comparison of memory discrimination with different prompt types using the Balanced Interaction Dataset. \textbf{Bold} numbers represent the average ranking of different prompts across all models. \textcolor{red}{\textbf{Red}} numbers represent the highest rank. \\ $^\dagger$ indicates the use of a quantized model.}
\centering
\resizebox{\linewidth}{!}{
\begin{tabular}{@{}c|cccc|ccc|
>{\columncolor[HTML]{ECF4FF}}c 
>{\columncolor[HTML]{ECF4FF}}c 
>{\columncolor[HTML]{ECF4FF}}c |ccc@{}}
\toprule
                                 & \multicolumn{4}{c|}{\textbf{Direct Prompt}}             & \multicolumn{3}{c|}{\textbf{General Prompt}} & \multicolumn{3}{c|}{\cellcolor[HTML]{ECF4FF}\textbf{Cluster Prompt (Ours)}}                                     & \multicolumn{3}{c}{\textbf{Specific Prompt}} \\
\multirow{-2}{*}{\textbf{Model}} & \textbf{Acc} & \bm{$R_0$} & \bm{$R_1$}   & \bm{$wF_1$}  & \textbf{Acc}  & \bm{$R_1$}    & \bm{$wF_1$}  & \textbf{Acc}                        & \bm{$R_1$}                          & \bm{$wF_1$}                         & \textbf{Acc}  & \bm{$R_1$}    & \bm{$wF_1$}  \\ \midrule
MiniCPM-2B                       & 54.24        & 31.73      & 81.13        & 51.57        & 59.74         & 97.83         & 50.63        & 78.40                               & 95.33                               & 77.43                               & 72.05         & 95.46         & 69.78        \\ \midrule
Qwen2-0.5B                       & 46.96        & 92.87      & 8.39         & 36.07        & 48.71         & 87.26         & 41.79        & 65.70                               & 94.95                               & 63.49                               & 65.69         & 95.06         & 63.48        \\
Qwen2-1.5B                       & 54.34        & 0.01       & 99.99        & 38.27        & 55.40         & 94.50         & 44.84        & 70.91                               & 77.84                               & 70.70                               & 70.80         & 81.93         & 70.27        \\
Qwen2-7B                         & 55.49        & 75.83      & 38.41        & 54.10        & 64.49         & 95.92         & 59.26        & 85.42                               & 89.47                               & 85.36                               & 82.11         & 85.48         & 82.07        \\
Qwen2-72B$^\dagger$              & 50.61        & 90.73      & 17.02        & 43.38        & 58.63         & 80.99         & 57.09        & 86.28                               & 84.19                               & 86.31                               & 83.25         & 84.33         & 83.26        \\ \midrule
LLAMA3-8B                        & 62.19        & 66.40      & 58.66        & 62.24        & 64.90         & 90.49         & 61.60        & 85.09                               & 92.08                               & 84.95                               & 82.76         & 91.54         & 82.52        \\
LLAMA3.1-8B                      & 62.37        & 43.48      & 78.12        & 61.13        & 61.29         & 95.86         & 54.25        & 85.86                               & 83.79                               & 85.89                               & 81.85         & 87.34         & 81.74        \\
LLAMA3-70B$^\dagger$             & 50.12        & 87.44      & 18.77        & 43.88        & 49.67         & 79.96         & 45.72        & 85.03                               & 82.65                               & 85.06                               & 81.56         & 79.99         & 81.60        \\ \midrule
ChatGLM3-6B                      & 53.36        & 76.16      & 34.22        & 51.44        & 56.22         & 86.35         & 50.63        & 70.48                               & 95.12                               & 67.83                               & 70.97         & 97.74         & 67.86        \\
GLM4-9B                          & 59.46        & 58.40      & 60.35        & 59.52        & 64.07         & 69.01         & 63.99        & 83.61                               & 95.87                               & 83.18                               & 81.82         & 94.83         & 81.31        \\ \midrule
\textbf{Average Rank}            & \textbf{3.8} & \textbf{-} & \textbf{3.7} & \textbf{3.6} & \textbf{3.2}  & \textbf{2.3}  & \textbf{3.4} & {\color[HTML]{FE0000} \textbf{1.1}} & {\color[HTML]{FE0000} \textbf{1.9}} & {\color[HTML]{FE0000} \textbf{1.1}} & \textbf{1.9}  & \textbf{2.1}  & \textbf{1.9} \\ \bottomrule
\end{tabular}}
\label{tab:prompt-compare}
\end{table}

\subsubsection{Ablation Study}

\paragraph{Effectiveness of Gating Unit Module.} The results in the middle part of \autoref{tab:total} show that GUM significantly reduces time costs while improving accuracy. Especially under real data distributions, it achieves nearly three times the speed improvement. This shows that the proposed identifier and multi-view role-playing quickly identify non-memorable and memorable data, respectively.

\begin{figure}[pos=t]
    \centering
    \includegraphics[width=5in]{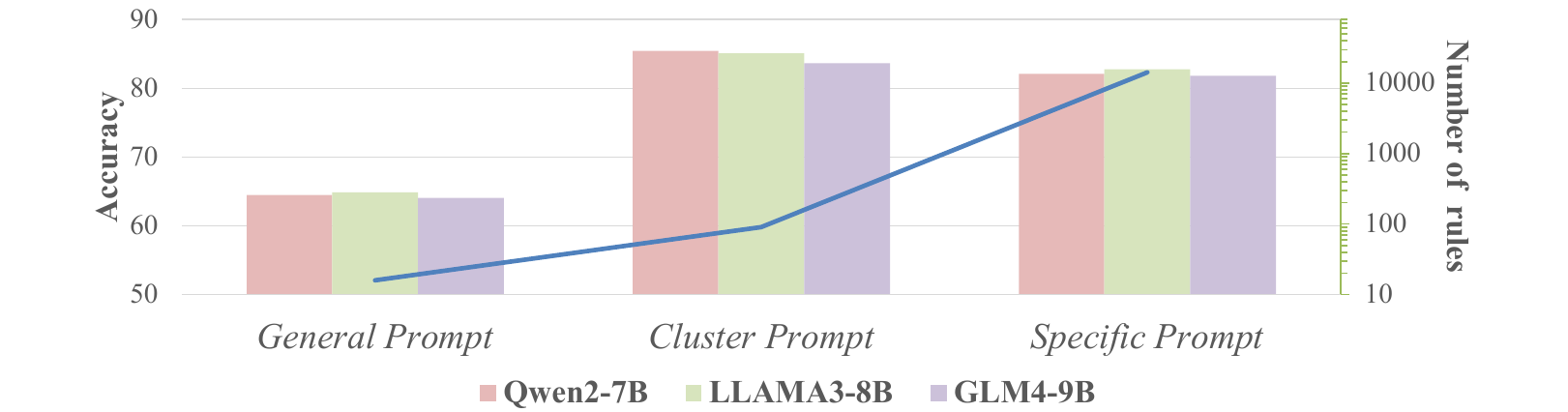}
    \caption{Comparison of different prompt types. The proposed cluster prompt, using only a similar magnitude of rules as the general prompt, performs as well or even better than the specific prompt.}
    \label{fig:prompt-compare}
\end{figure}

\paragraph{Effectiveness of Cluster Prompting Module.}
A thorough examination of the results in \autoref{tab:prompt-compare} and \autoref{fig:prompt-compare} allows us to investigate the impact of different prompts on memory discrimination. We conduct experiments under the following four settings:

\begin{itemize}
\item \textbf{Direct Prompt}: Directly ask LLMs, `\textit{Does the text contain personal information that needs recording?}'. Different LLMs all hit around 50\% accuracy. However, the differences in the performance of recall reveal the inherent biases of each LLM on what to remember. In truth, LLMs lack real memory understanding and cannot be used for memory discrimination directly.

\item \textbf{General Prompt}: Inform LLMs of criteria for judging what content to remember based on the personal information required by applications through a set of general rules. These rules include that the content needs to involve aspects such as personal education, work, health, social relationships, and travel activities. The results indicate that this increases the recall of memorable data to over 80\% while having little impact on accuracy. LLMs still do not understand what not to remember.

\item \textbf{Specific Prompt}: Create several discrimination rules for each intent based on all memory scenes to enhance LLMs' understanding of memory. The results indicate that the performance of memory discrimination has improved by nearly 20\%. Nevertheless, it depends heavily on thousands of manually crafted rules and the system's accurate intent recognition results. Additionally, rules designed for a single intent might not cover all user behaviors, potentially leading to gaps in real scenarios.

\item \textbf{Cluster Prompt}: The proposed cluster prompt groups intents covering similar memory scenes. We construct positive and negative memory discrimination rules for each group. The results show that SAMD achieves optimal performance on most LLMs. Compared to the general prompt, SAMD uses a similar number of rules but improves performance by over 20\%. Compared to the specific prompt, SAMD uses less than 1\% of the number of rules while outperforming them.
\end{itemize}

\paragraph{Effectiveness of Intent-Scene Joint Clustering.}

The results are shown in \autoref{tab:cluster}. 
Intent embeddings with density-based clustering algorithms are ineffective due to the inherent vagueness and ambiguity of intent names.
Then, we further incorporate relevant interactions for each intent in embeddings, which improves intent understanding and the discriminative power of the embeddings, though only slightly.
Significant manual adjustments remained necessary before usage.
The proposed intent-scene joint clustering effectively captures intent correlations by creating the affinity matrix between intents and memory scenes. By analyzing latent intent aggregations via SVD, we achieve clustering that aligns with actual semantic meanings for subsequent clustering prompt construction.

\begin{table}[t]
\caption{Quantitative results for intent clustering are assessed in terms of both clustering and multi-class classification. The \textbf{bold} number highlights the best results. K-Means represents the Bisecting K-Means algorithm.}
\centering
\resizebox{\linewidth}{!}{
\begin{tabular}{@{}c|cccccc|cccccc|c@{}}
\toprule
\multirow{2}{*}{\textbf{Metrics}} & \multicolumn{3}{c|}{\textbf{BERT}} & \multicolumn{3}{c|}{\textbf{GPT-2}} & \multicolumn{3}{c|}{\textbf{BERT}} & \multicolumn{3}{c|}{\textbf{GPT-2}} & \multirow{2}{*}{\textbf{Ours}} \\
 & \multicolumn{1}{l}{\textbf{Meanshift}} & \multicolumn{1}{l}{\textbf{K-Means}} & \multicolumn{1}{l|}{\textbf{Spectral}} & \multicolumn{1}{l}{\textbf{Meanshift}} & \multicolumn{1}{l}{\textbf{K-Means}} & \multicolumn{1}{l|}{\textbf{Spectral}} & \multicolumn{1}{l}{\textbf{Meanshift}} & \multicolumn{1}{l}{\textbf{K-Means}} & \multicolumn{1}{l|}{\textbf{Spectral}} & \multicolumn{1}{l}{\textbf{Meanshift}} & \multicolumn{1}{l}{\textbf{K-Means}} & \multicolumn{1}{l|}{\textbf{Spectral}} &  \\ \midrule
\textbf{$RI$} & 0.029 & 0.033 & \multicolumn{1}{c|}{0.086} & 0.009 & 0.298 & 0.037 & 0.071 & 0.138 & \multicolumn{1}{c|}{0.107} & 0.016 & 0.333 & 0.081 & \textbf{0.626} \\
\textbf{$wF_1$} & 0.05 & 0.24 & \multicolumn{1}{c|}{0.24} & 0.04 & 0.33 & 0.20 & 0.09 & 0.25 & \multicolumn{1}{c|}{0.29} & 0.06 & 0.38 & 0.30 & \textbf{0.71} \\ \midrule
\multicolumn{1}{c|}{-} & \multicolumn{6}{c|}{\textit{Embedding: Intent Only}} & \multicolumn{6}{c|}{\textit{Embedding: Intent with corresponding texts}} & \multicolumn{1}{l}{} \\ \bottomrule
\end{tabular}}
\label{tab:cluster}
\end{table}

\subsection{Indirect Evaluation of Memory Discrimination}
\label{sec:indirect}

We indirectly evaluate memory discrimination by assessing its impact on end-to-end memory agents, verifying that improved memory enhances application performance and reduces cost.
The successful usage reflects the utility of the memory discrimination method. We experiment on the state-of-the-art methods that store raw data and summaries, Self-Controlled Memory framework (SCM)~(\cite{wang2024enhancinglargelanguagemodel}), and MemoryBank~(\cite{DBLP:conf/aaai/ZhongGGYW24}).

\subsubsection{Experimental Settings}

\paragraph{Datasets.} We choose two widely used datasets to evaluate the memory-assisted agents, and the conversation data is a typical form of personal interaction.
\begin{itemize}
    \item \textbf{Dataset in SCM}: Contains 18 dialogue instances collected from the open source data ShareChat\footnote{https://paratranz.cn/projects/6725} and 105 human-annotated probing questions. 
    \item \textbf{Dataset in MemoryBank}:  Comprises conversations from 15 virtual users over 10 days and has a total of 194 questions. These users have diverse personalities, and the dialogue on each day covers at least two topics.
\end{itemize}

\paragraph{Metrics.} Following SCM and MemoryBank, we evaluate across multiple dimensions, including answer accuracy, efficiency, and retrieval accuracy:
\begin{itemize}
    \item \textbf{Answer accuracy}: Evaluate the accuracy of answers to probing questions. Note that we further evaluate questions that require single-turn dialogue and multi-turn dialogue history separately for SCM.
    \item \textbf{Computational cost}: Evaluate the average number of tokens on each dialogue. Since all methods rely on LLM APIs, under the same network conditions, this directly reflects the efficiency and runtime of these methods.
    \item \textbf{Memory retrieval effect}: Determines if related memory can be successfully retrieved. Specifically, we calculate the recall and Top-1 accuracy of the retrieved memory.
\end{itemize}

\paragraph{Implementation Details.} For fair comparison, we perform experiments using the open-source code\footnote{https://github.com/wbbeyourself/SCM4LLMs}\footnote{https://github.com/zhongwanjun/MemoryBank-SiliconFriend}.

\begin{table}[t]
\caption{The comparison results of SCM with and without the proposed SAMD. Referring to the original paper, we conduct experiments on both \texttt{GPT-3.5-turbo} and \texttt{GPT-4o-mini}. We apply the default settings of SCM and utilize the Top-4 retrieved memory. The \textcolor{red}{red} numbers indicate an increase in performance, while the \textcolor{darkgreen}{green} numbers indicate a decrease in consumption. SAMD helps construct superior memory, resulting in more accurate question answering, improved memory retrieval accuracy, and reduced computational costs.}
\centering
\resizebox{\linewidth}{!}{
\begin{tabular}{@{}l|ccc|c|cc@{}}
\toprule
                       \multirow{2}{*}{\textbf{Method}} & \multicolumn{3}{c|}{\textbf{Answer Accuracy}}                           & \multicolumn{1}{l|}{\multirow{2}{*}{\textbf{Avg. Tokens/Dialogue}}} & \multicolumn{2}{c}{\textbf{Memory Retrieval}}  \\
                       & Total                  & Single Turn           & Multi Turn             & \multicolumn{1}{l|}{}                                               & Recall                & Top-1 Accuracy             \\ \midrule
SCM$_\text{turbo}$    & 69.2                   & 77.2                  & 63.3                   & 86k                                                                 & 84.1                  & 57.7                   \\
+ \textbf{SAMD}  & \textbf{82.6 (\textcolor{red}{+13.4\%})} & \textbf{77.2 (\textcolor{gray}{+0.0\%})} & \textbf{86.7 (\textcolor{red}{+23.4\%})} & \textbf{64k (\textcolor{darkgreen}{-25.6\%})}                                               & \textbf{88.5 (\textcolor{red}{+4.4\%})} & \textbf{75.0 (\textcolor{red}{+17.3\%})} \\ \midrule
SCM$_\text{4o-mini}$ & 65.4                   & 72.7                  & 60.0                   & 91k                                                                 & 81.4                  & 59.6                   \\
+ \textbf{SAMD}  & \textbf{80.0 (\textcolor{red}{+14.6\%})} & \textbf{77.2 (\textcolor{red}{+4.5\%})} & \textbf{83.3 (\textcolor{red}{+23.3\%})} & \textbf{65k (\textcolor{darkgreen}{-28.6\%})}                                               & \textbf{85.0 (\textcolor{red}{+3.6\%})} & \textbf{71.2 (\textcolor{red}{+11.6\%})} \\ \bottomrule
\end{tabular}}
\label{tab:scm}
\end{table}

\begin{table}[t]
\caption{The comparison results of \texttt{ChatGPT}-based SiliconFriend (MemoryBank) with and without SAMD. The introduction of memory discrimination significantly reduces computational costs, enhances retrieval accuracy, and improves the accuracy.}
\centering
\resizebox{3.5in}{!}{
\begin{tabular}{@{}l|cc|c@{}}
\toprule
                                        \textbf{Method} & \textbf{Answer Acc.} & \textbf{Retrieval Acc.} & \textbf{Avg. Token} \\ \midrule
SiliconFriend & 0.73                 & 0.79                    & 16.5k                      \\
+ \textbf{SAMD}                   & \textbf{0.78 (\textcolor{red}{+5\%})}  & \textbf{0.85 (\textcolor{red}{+6\%})}     & \textbf{7.4k (\textcolor{darkgreen}{-55.2\%})}     \\ \bottomrule
\end{tabular}}
\label{tab:memorybank}
\end{table}

\subsubsection{Quantitative Results}

\autoref{tab:scm} presents the SCM results, demonstrating that incorporating memory discrimination reduces token consumption by over 25\%. SAMD filters out sentences in the context that are irrelevant to personal information or meaningless, resulting in a high-quality memory, which decreases storage needs and invalid inputs for other memory operations. Recall of historical dialogues related to questions has improved, with at least a 10\% boost in Top-1 retrieval accuracy. Consequently, answer accuracy improves by over 10\% and over 20\% for multi-turn questions, thanks to retrieval results that contain more relevant information and less noise, enhancing overall information coverage.
\autoref{tab:memorybank} provides the MemoryBank results. The memory discrimination method SAMD helps avoid more than 50\% of unnecessary computational costs. With reduced noise in memory, the accuracy of retrieval historical information and responses improves by at least 5\%.

\begin{figure}[pos=t]
    \centering
    \includegraphics[width=\linewidth]{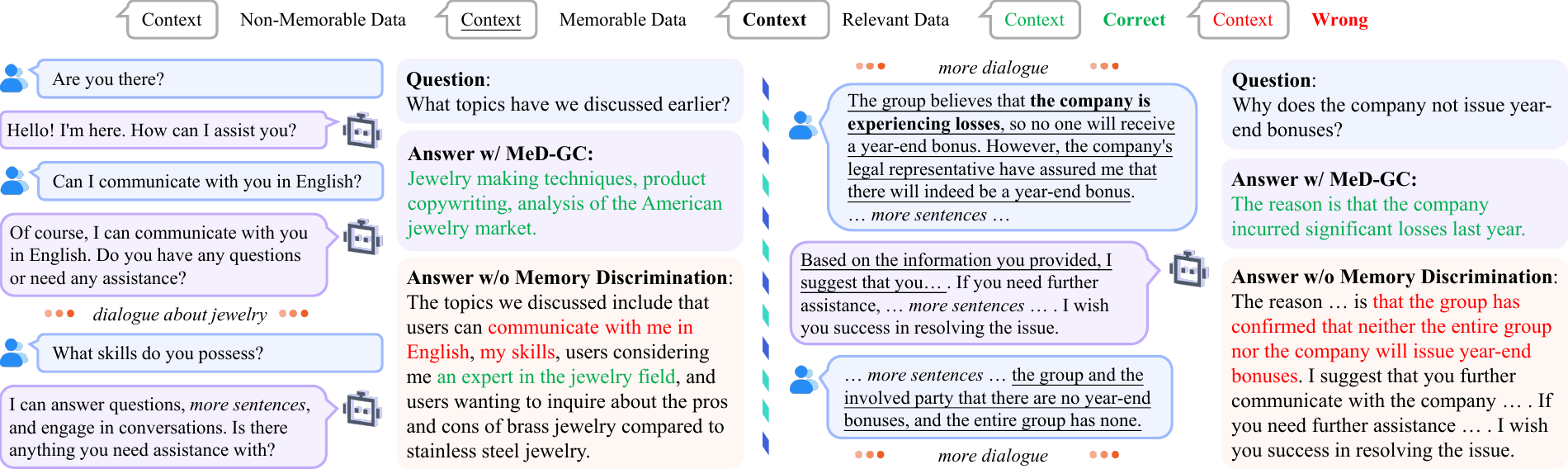}
    \caption{Examples of the probing questions in SCM's data. There is a large amount of non-memorable content. As indicated on the left side, memory discrimination can prevent the introduction of noise in the answer and the loss of answer elements. On the right side, memory discrimination helps avoid relevant information being drowned in a plethora of non-memorable content, allowing it to be correctly retrieved and utilized.}
    \label{fig:scm}
\end{figure}

\begin{figure}[pos=t]
    \centering
    \includegraphics[width=\linewidth]{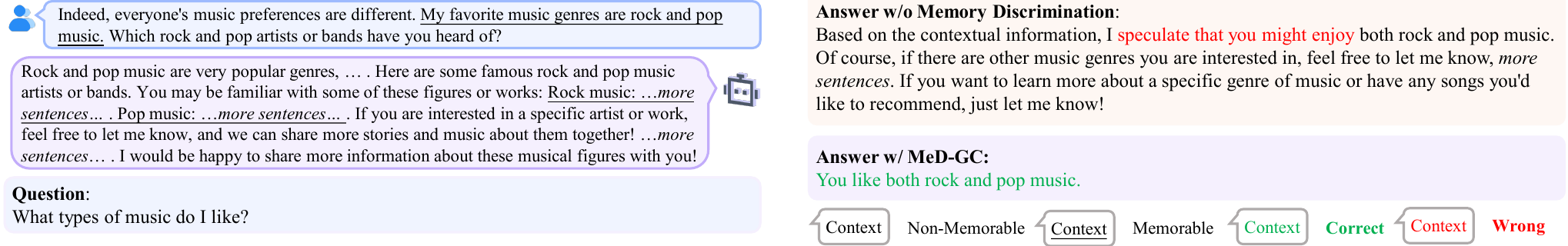}
    \caption{Example of probing questions in MemoryBank's data. The storage of excessive non-memorable data in memory causes LLMs to have biases in understanding them during use, leading to incorrect responses. Additionally, it makes the responses verbose.}
    \label{fig:memorybank}
\end{figure}

\subsubsection{Qualitative Results}

We find that users' daily interactions contain a large amount of meaningless data. As shown on the left side of \autoref{fig:scm}, if these data are incorrectly activated during application, it causes inaccurate responses and omissions of relevant information. Moreover, as shown on the right side, when the amount of context gradually increases, the lack of memory discrimination makes LLMs overlook key clues, resulting in an incomplete understanding of events. This leads to incorrect conclusions in responses. The proposed SAMD effectively filters out non-memorable data, leading to concise and accurate responses. 
As seen in \autoref{fig:memorybank}, without memory discrimination, meaningless information interferes with LLMs' understanding of user personal information. This causes neglect or incorrect inferences about user information. Moreover, when this low-quality memory is used as a reference, we receive verbose and inaccurate responses. However, with the introduction of SAMD, we obtain concise and accurate ones.

\subsubsection{Failure Case Study}

Analysis of incorrect responses reveals that while historical information is generally stored accurately, memory-assisted agents struggle due to limitations in other memory modules. We identify two primary issues: incomplete responses and misinterpretation of questions.
Firstly, when too many memory fragments are required, some may not be activated properly. This results in incomplete or inaccurate responses. For example, when a user asks,    
\begin{center}
    \textit{``What topics did we discuss?"}, 
\end{center}
and there is a large amount of content under each topic. Some topics may be lost in the response due to retrieval limitations. Secondly, questions are sometimes not correctly understood. For example, when a user asks,
\begin{center}
    \textit{``What is the first poem I sent you?"}.
\end{center}
Memory-assisted agents may struggle to perform operations in memory to locate the `first' item. Therefore, memory discrimination streamlines memory, providing a more concise and efficient personal information foundation for memory writing, management, and reading modules, thus unlocking tremendous potential for memory-assisted agents.

\section{Conclusion}
\label{sec:conclusion}

In this paper, we propose a scene-aware memory discrimination method via GUM and CPM called SAMD, aimed at enhancing the efficiency and quality of memory construction for personal daily interaction data. SAMD effectively recalls memorable data while maintaining robustness in complex or changing scenarios. SAMD not only provides new insights for personal memory processing but also lays the groundwork for advancing the application of intelligent devices in personalized services. We look forward to further exploration in this field to achieve more efficient memory management and a better user experience. \\

\noindent \textbf{Acknowledgement.}
This work was supported by the National Natural Science Foundation of China (U23B2057), and Shanghai Pilot Program for Basic Research (22TQ1400300).

% To print the credit authorship contribution details
\printcredits

%% Loading bibliography style file
\bibliographystyle{model1-num-names}
% \bibliographystyle{cas-model2-names}

% Loading bibliography database
\bibliography{cas-refs}

% Biography
%\bio{}
% Here goes the biography details.
%\endbio

%\bio{pic1}
% Here goes the biography details.
%\endbio

\end{document}